\documentclass[10pt,twocolumn]{article}

\usepackage[preprint]{neurips_2023}
\usepackage[T1]{fontenc}
\usepackage[utf8]{inputenc}
\usepackage{microtype}
\usepackage{amsmath,amssymb}
\usepackage{array}
\usepackage{booktabs}
\usepackage{caption}
\usepackage{enumitem}
\usepackage{graphicx}
\usepackage{hyperref}
\usepackage{multirow}
\usepackage{placeins}
\usepackage{tabularx}
\usepackage{xspace}

\graphicspath{{figures/}}
\hypersetup{
  colorlinks=true,
  linkcolor=black,
  citecolor=black,
  urlcolor=black,
  filecolor=black
}
\newcommand{\method}{\textsc{Write-Subspace LoRA}\xspace}
\newcommand{\pcasub}{\textsc{PCA-Subspace}\xspace}
\newcommand{\randomsub}{\textsc{Random-Subspace}\xspace}
\newcommand{\probesub}{\textsc{Probe-Subspace}\xspace}
\newcommand{\dora}{\textsc{DoRA}\xspace}
\newcommand{\pissa}{\textsc{PiSSA}\xspace}

\newcommand{\mainmethod}{\textsc{Learned-Basis LoRA}\xspace}
\newcommand{\full}{\textsc{Full}\xspace}
\newcommand{\code}[1]{\texttt{#1}}
\newcolumntype{L}[1]{>{\raggedright\arraybackslash}p{#1}}

\makeatletter
\renewenvironment{abstract}
  {%
    \begin{center}\large\textbf{\abstractname}\end{center}%
    \begin{list}{}%
      {%
        \setlength{\rightmargin}{0.6cm}%
        \setlength{\leftmargin}{0.6cm}%
      }%
      \item[]\ignorespaces%
      \@setsize\normalsize{12pt}\xpt\@xpt
  }
  {%
    \unskip\end{list}%
  }
\renewcommand{\maketitle}{%
  \par
  \begingroup
    \renewcommand{\thefootnote}{\fnsymbol{footnote}}%
    \renewcommand{\@makefnmark}{\hbox to \z@{$^{\@thefnmark}$\hss}}%
    \long\def\@makefntext##1{%
      \parindent 1em\noindent
      \hbox to 1.8em{\hss $\m@th ^{\@thefnmark}$}##1
    }%
    \if@twocolumn
      \twocolumn[\@maketitle]%
    \else
      \thispagestyle{empty}%
      \@maketitle%
    \fi
    \@thanks
    \@notice
  \endgroup
  \let\maketitle\relax
  \let\thanks\relax
}
\makeatother

\title{Behaviorally Effective LoRA Writes Are Sparse and Structured}
\author{
Haruto Sato, Yuki Tanaka, Ren Nakamura,\\
Aoi Kobayashi, Mei Ito\\
Independent Researchers
}

\begin{document}

\maketitle
\begin{abstract}
Low-rank adaptation fixes the rank of the update, but it does not identify which parts of a trained write actually carry behavior. We study that question directly and show that behaviorally effective LoRA writes are sparse, structured, and far more concentrated than the raw low-rank parameterization suggests.

We use \mainmethod, a learned-basis continuation recipe, to expose that structure. The recipe warms up an unconstrained adapter, converts its learned write columns into a module-wise orthonormal basis, freezes that basis, and continues training inside the constrained parameterization. Across 14 exact switches from unconstrained to constrained form, held-out accuracy is unchanged at the conversion step and reconstructed write matrices differ by at most 0.25\% relative Frobenius error. Same-state continuation then shows that the same trained checkpoint develops differently under different write subspaces, establishing write geometry as a causal state variable. A no-retraining projection test shows that useful write signal stays inside the learned write space and largely disappears from random or frozen-activation PCA controls.

The concentration pattern is strong at both local and global scales. Across GSM8K, MathQA, and AQuA, per-module top-\(k\) continuation reaches its optimum at \(k \in \{2,4\}\) in all twelve seed-level cases we test. A stricter global ranking test shows that learned top-16 and top-32 subsets outperform matched random subsets, especially on GSM8K/Qwen and MathQA/Qwen. Single-direction ablations further reveal a sparse set of late \code{q\_proj}, \code{o\_proj}, and \code{down\_proj} components with outsized behavioral impact.

The contribution of the paper is a structural mechanism result. Trained LoRA adapters do not use their write budget uniformly. Most of the downstream effect is carried by a small, late, and structured subset of write components, and a learned-basis constrained continuation exposes that organization clearly enough to measure it.
\end{abstract}

\section{Introduction}

Parameter-efficient fine-tuning has become the standard way to adapt large language models under strict memory, storage, and optimizer budgets \citep{houlsby2019adapters,li2021prefix,lester2021power,liu2021ptuningv2,zaken2022bitfit,karimi2021compacter,hu2021lora,dettmers2023qlora,ding2023peftsurvey,zhang2023adalora,kopiczko2023vera}. Most PEFT work studies scalar capacity decisions: rank, parameter budget, quantization level, or layer allocation. A growing line instead studies update geometry through weight decomposition, singular-structure initialization, orthogonal constraints, and structured subspaces \citep{liu2024dora,meng2024pissa,wang2024milora,she2025dislora,liu2024boft,wang2023olora,lion2025polar,li2025stella,yang2025srlora,xiao2026structlora,chang2026soslora}. These papers show that geometry matters. They leave a sharper question open: after a LoRA adapter has been trained, which parts of its write update actually carry task behavior?

Recent systems in adjacent areas increasingly treat model capacity as a selectively allocated resource. EXACT redistributes long-context supervision toward underexposed long-range targets \citep{zhu2026exact}. HeLa-Mem consolidates agent experience through associative memory structure \citep{zhu2026helamem}. GapSight learns when a vision-language model should revisit a local visual region \citep{zhu2026lookagain}. FCPRAG performs retrieval-conditioned fusion over passage-specific LoRA adapters \citep{zhu2026fcprag}. Learning Less Is More shows that premature upper-layer attention specialization can distort pretraining dynamics before lower-layer features stabilize \citep{zhu2026learningless}. The shared pattern is selective allocation: the decisive signal occupies only part of the available computation or parameter space. We ask whether LoRA writes follow the same pattern.

A standard LoRA adapter can spend its limited rank on any output direction that improves training loss, including prompt-format quirks, split-specific shortcuts, or brittle heuristics. If transfer depends on only a small part of that update, rank alone is a poor description of what the adapter learned. We therefore separate two objects: the low-rank code the adapter computes and the output subspace into which it writes. The central claim of this paper is direct: behaviorally effective LoRA writes are sparse and structured.

We study this with a projected low-rank parameterization that freezes a module-wise write basis and continues training inside that subspace. Our main trainable instantiation, \mainmethod, first warms up an unconstrained \full adapter, orthonormalizes its learned write columns into a basis, and then continues training with the basis fixed. This construction lets us run four clean tests. Exact conversion verifies that the switch into constrained form is lossless. Same-state basis swap tests whether write geometry is a causal state variable. No-retraining projection tests where the useful signal lies inside a trained \full update. Local and global concentration analyses measure how much of the learned write space is actually needed.

The evidence supports the same conclusion across several intervention levels. Across 14 exact switches, held-out accuracy is unchanged at the conversion step and reconstructed write matrices differ by at most 0.25\% relative Frobenius error. Same-state continuation from a shared checkpoint shows large separations among candidate bases, with the learned basis best in every benchmark-backbone pair we test. No-retraining projection preserves almost all \full performance inside the learned component and removes it from random or frozen-activation PCA controls. Across GSM8K, MathQA, and AQuA, the best per-module continuation always appears at \(k \in \{2,4\}\) across twelve seed-level cases. Global top-16 and top-32 learned subsets also outperform matched random subsets, and the largest single-direction effects cluster in late \code{q\_proj}, \code{o\_proj}, and \code{down\_proj} modules.

The main text centers on real-task mechanism evidence. The synthetic graph benchmark stays in the appendix, where it isolates shortcut shift and structural length shift under controlled conditions. Its role is to explain why write geometry can matter once the real-task diagnostics establish that the constrained parameterization is viable.

Our contributions are:
\begin{enumerate}[leftmargin=1.5em]
\item We identify write geometry as a causal state variable of PEFT: from the same trained checkpoint, different write subspaces lead to different futures.
\item We show that trained LoRA writes are sharply concentrated: per-module optima appear at top-2 or top-4 learned directions, and learned global subsets outperform matched random subsets.
\item We localize the strongest effects to a sparse set of late \code{q\_proj}, \code{o\_proj}, and \code{down\_proj} components, and we use \mainmethod as a mechanism probe that exposes that organization on real reasoning benchmarks.
\end{enumerate}

\section{Related Work}

\subsection{PEFT and Low-Rank Adaptation}

PEFT methods reduce adaptation cost by freezing most backbone parameters and learning only a small task-specific interface. Early adapter work showed that small residual modules can recover much of the benefit of full fine-tuning \citep{houlsby2019adapters}. Later variants broadened that idea through prompt tuning, prefix tuning, P-Tuning v2, BitFit, Compacter, and LoRA-style low-rank updates \citep{li2021prefix,lester2021power,liu2021ptuningv2,zaken2022bitfit,karimi2021compacter,hu2021lora}. QLoRA, AdaLoRA, and VeRA further optimized that interface for quantization, budget allocation, and storage efficiency \citep{dettmers2023qlora,zhang2023adalora,kopiczko2023vera}. This literature establishes low-rank adaptation as an efficient training interface. Our paper keeps that background fixed and asks a more structural question: once a strong task-specific LoRA update has been learned, how much of it is actually behaviorally essential?

\subsection{Geometry-Aware and Spectral Variants of LoRA}

The closest thread is the growing line of geometry-aware LoRA variants. Some methods change the parameterization itself, for example by separating direction and magnitude as in \citet{liu2024dora}. Others initialize LoRA from spectral structure in pretrained weights, such as \citet{meng2024pissa}, or emphasize different parts of that structure, such as \citet{wang2024milora}. A third group explicitly constrains or learns subspaces through orthogonal, butterfly, or manifold structure, including \citet{she2025dislora}, \citet{liu2024boft}, \citet{wang2023olora}, \citet{lion2025polar}, and \citet{li2025stella}. More recent work updates or diversifies subspaces during training, as in \citet{yang2025srlora}, \citet{chang2026soslora}, and \citet{xiao2026structlora}. These papers are highly relevant because they show that update geometry is a first-order design choice rather than an incidental implementation detail.

The paper uses a geometry-constrained adapter as an instrument to expose the internal structure of the trained write update. The central question is: once a useful write space has formed, how concentrated is the behaviorally effective part of the update inside that space? This is why the core experiments are no-retraining projection, same-state basis-swap continuation, and top-\(k\) continuation inside the learned basis.

Recent post-hoc work is even closer in spirit. LoRA-Squeeze learns a larger adapter and compresses it into a smaller one \citep{vulic2026lorasqueeze}. Spectral Surgery reweights singular directions inside a trained adapter while keeping the learned subspace fixed \citep{tian2026spectralsurgery}. Both methods assume that trained adapters contain internal redundancy. Our paper supplies direct behavioral evidence for that assumption and identifies where the effective write signal is concentrated.

\subsection{Low-Dimensional Fine-Tuning and Representation Subspaces}

Several results suggest that fine-tuning behaves as a low-dimensional phenomenon \citep{aghajanyan2020intrinsic}. That observation helps explain why low-rank or sparse parameterizations can work at all, but it does not by itself identify which low-dimensional directions matter for behavior. Our method is also related to representation-level intervention methods such as \citet{wu2024reft}, which edit or finetune behavior through low-dimensional hidden-state interventions, and to projection-based techniques such as iterative nullspace projection and concept erasure \citep{ravfogel2020null,belrose2023leace}. The conceptual overlap is that subspaces in hidden space can have semantic or functional significance. The difference is that we use this idea as a persistent training-time write constraint inside LoRA and then ask a behavior-level question: how many learned write directions are actually needed for the best continuation?

\subsection{Structured Control, Routing, and Selective Allocation}

The broader representation-engineering literature shows that linear directions in hidden space can strongly influence behavior without modifying the full model weights \citep{subramani2022extracting,turner2023actadd,zou2023repe,meng2022rome}. Adapter composition and routing work makes a related point in parameter space: MoLE and MixLoRA combine multiple LoRA experts through learned control \citep{wu2024mole,li2024mixlora}, and FCPRAG performs sample-level fusion over passage-specific LoRA injections \citep{zhu2026fcprag}. Recent work beyond standard PEFT applies the same selective-allocation principle in other forms. EXACT redistributes long-context supervision \citep{zhu2026exact}, HeLa-Mem consolidates agent memory through associative structure \citep{zhu2026helamem}, GapSight learns when a vision-language model should revisit local evidence \citep{zhu2026lookagain}, and Learning Less Is More shows that upper-layer attention timing can hurt pretraining when specialization outruns lower-layer feature formation \citep{zhu2026learningless}. These papers solve different tasks, but they share the idea that useful signal is localized and should be routed, weighted, or constrained selectively.

\subsection{Where the Present Claim Sits}

Taken together, these prior works already imply that geometry matters in adaptation. Our claim sits one step downstream. We ask what fraction of a trained write update is behaviorally essential once geometry has already been learned. The answer supported by our interventions is that the task-effective part of a trained LoRA write update is sparse and structured. Exact conversion, no-retraining projection, same-state basis-swap continuation, local top-\(k\) continuation, and global top-\(M\) selection all point to that same conclusion.

\section{Method}

\subsection{Constrained Write Subspaces}

Let $h_\ell \in \mathbb{R}^{d}$ be a hidden state at layer $\ell$. Standard LoRA adds a low-rank update
\begin{equation}
\Delta h_\ell = B_\ell A_\ell h_\ell,
\end{equation}
where $A_\ell \in \mathbb{R}^{r \times d}$ and $B_\ell \in \mathbb{R}^{d \times r}$. This constrains rank, but it leaves the output directions of the write unconstrained.

We separate two questions:
\begin{enumerate}[leftmargin=1.5em]
\item What is the low-rank code the adapter computes?
\item In which output subspace is that code allowed to write?
\end{enumerate}
To make the second question explicit, we fix a module-wise orthonormal basis $U_\ell \in \mathbb{R}^{d \times k}$ with $U_\ell^\top U_\ell = I$ and parameterize
\begin{equation}
\Delta h_\ell = U_\ell C_\ell A_\ell h_\ell,
\end{equation}
where $C_\ell \in \mathbb{R}^{k \times r}$ is learned and $U_\ell$ is frozen. Every adapter write is therefore confined to $\mathrm{span}(U_\ell)$. We refer to this parameterization as \method.

\subsection{Frozen-Basis Baselines}

We study two direct ways to choose $U_\ell$ before training.

\paragraph{Random basis.}
Sample a Gaussian matrix in $\mathbb{R}^{d \times k}$ and orthonormalize it. This baseline keeps the same bottleneck size but removes any geometry signal.

\paragraph{Frozen-activation PCA basis.}
Collect frozen activations on the train split, form the activation matrix $H_\ell \in \mathbb{R}^{n \times d}$ for each target layer, center it, and keep its top-$k$ right singular vectors. This basis reflects directions already used by the frozen backbone on the task distribution, but it is estimated before adaptation begins.

These baselines are scientifically useful because they answer whether any generic bottleneck helps and whether a purely frozen representation geometry is already sufficient.

\subsection{Warmup-Induced Basis Extraction}

Instead of choosing $U_\ell$ from the frozen backbone alone, \mainmethod estimates it from a partially trained \full adapter. Suppose a warmup stage has produced a target-module write matrix $B_\ell^\text{warmup}$. We then set
\begin{equation}
U_\ell = \mathrm{orth}(B_\ell^\text{warmup}), \qquad C_\ell = U_\ell^\top B_\ell^\text{warmup},
\end{equation}
where $\mathrm{orth}(B_\ell^\text{warmup})$ is an orthonormal basis for the column space of the warmup write. Then
\begin{equation}
B_\ell^\text{warmup} = U_\ell C_\ell,
\end{equation}
so the warmup update can be rewritten exactly as a constrained update:
\begin{equation}
B_\ell^\text{warmup} A_\ell h_\ell = U_\ell C_\ell A_\ell h_\ell.
\end{equation}

This conversion matters because it lets training continue from the warmup solution without changing the represented update at the switch step. Any later gain or loss must therefore come from the frozen write space used after the switch.

\subsection{A Trainable Recipe}

Instead of estimating the basis entirely from the frozen backbone, we estimate it from a partially trained \full adapter.

\paragraph{Stage 1: warmup.}
Train an unconstrained \full adapter for $T_\text{warmup}$ steps.

\paragraph{Stage 2: freeze the learned write basis.}
For each target module, extract the orthonormal basis of the warmup adapter's learned output column space, convert the warmup solution into constrained form, and continue training only the constrained parameters.

This produces our main algorithm, \mainmethod. The idea is simple: learn a useful write space first, then freeze it and continue training inside it. The method interpolates between two extremes:
\begin{enumerate}[leftmargin=1.5em]
\item freezing the basis too early, before the warmup adapter has learned useful write directions;
\item never freezing the basis at all, which reduces to \full LoRA.
\end{enumerate}
The empirical question is what this recipe reveals about the internal organization of the trained write update.

\subsection{Why Timing and Module Alignment Matter}

Two implementation details are central in the real benchmark. First, the basis must be module-wise and layer-aligned. Reusing one basis across non-aligned adapter layers mixes incompatible write spaces and weakens continuation. Second, the basis must be frozen late enough to reflect a meaningful adaptation trajectory. The experiments below show that an early warmup can produce a basis that is exact for the current weak adapter and still poor for the final task, while a later warmup yields a much stronger constrained continuation.

This decomposition yields four falsifiable structural predictions. First, warmup-to-subspace conversion should be representationally continuous. Second, if write geometry is a causal state variable, the same trained checkpoint should have different futures under different write subspaces. Third, if the useful update is concentrated, only a very small number of learned directions should be needed inside each module, and a smaller-than-full learned subset should still dominate globally. Fourth, if the effect is structured, the strongest components should cluster in a limited set of late write modules. The experiments are organized around these predictions.

\subsection{Interpretation}

The paper studies a structural claim. For PEFT, rank leaves out an important variable: the directions into which the adapter is allowed to write. Those directions change held-out behavior, and the behaviorally effective part of the trained update is highly non-uniform. Random and frozen-activation PCA are frozen basis estimators; \mainmethod tests whether learning the basis during adaptation is enough to expose a sparse and structured subset of write components that carries most of the downstream effect.

\begin{figure*}[t]
\centering
\includegraphics[width=0.98\textwidth]{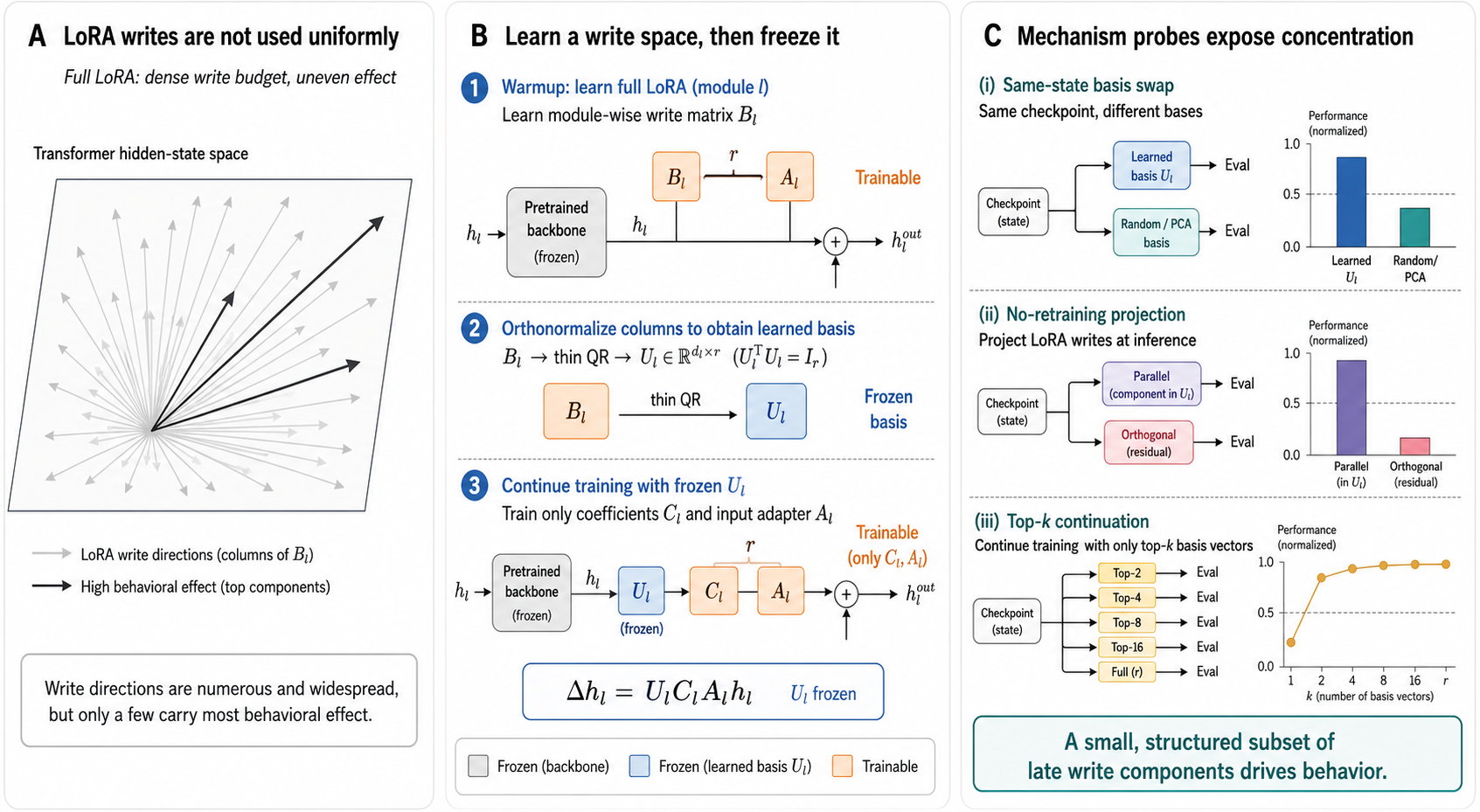}
\caption{Overview of the paper's mechanism pipeline. A trained LoRA adapter does not use its write budget uniformly. We warm up an unconstrained adapter, orthonormalize its module-wise write columns into a learned basis, freeze that basis, and continue optimization in constrained form. Same-state basis swap, no-retraining projection, and top-\(k\) continuation then test whether behavior is concentrated inside a small learned write subspace.}
\label{fig:method}
\end{figure*}

\section{Experimental Setup}

\subsection{Main Real Benchmarks}

The main text uses six partial real benchmarks: GSM8K-partial \citep{cobbe2021gsm8k}, CommonsenseQA \citep{talmor2019commonsenseqa}, StrategyQA \citep{geva2021strategyqa}, AQuA \citep{ling2017program}, ARC-Challenge \citep{clark2018arc}, and MathQA \citep{amini2019mathqa}. Each benchmark is evaluated on two 3B instruction-tuned backbones, Qwen2.5-3B-Instruct \citep{qwen2025technical} and Llama-3.2-3B-Instruct via the Llama 3 family \citep{dubey2024llama}. The main split uses 1,024 train examples, 128 validation examples, and up to 256 test examples for every benchmark-backbone pair; AQuA uses its full 254-example test split. All reported runs use seeds 39 and 40, the last four target layers, and the shared target-module family \code{q\_proj}, \code{o\_proj}, and \code{down\_proj}. We report held-out test accuracy in the main text and keep per-seed detail only for the timing diagnostic in the appendix.

The paper's primary evidence consists of four mechanism analyses: exact warmup-to-subspace conversion, same-state basis-swap continuation, no-retraining counterfactual projection, and concentration analysis via per-module top-\(k\), global top-\(M\), and single-direction ablations. The multi-benchmark table serves as scope evidence that the constrained parameterization remains viable beyond the intervention tests.

\subsection{Prompting and Readout}

All rows within a given benchmark share exactly the same prompt and readout rule. GSM8K uses a chain-of-thought prompt that requires a final sentence of the form ``The final answer is <number>.'' and is scored by direct numeric extraction. CommonsenseQA, AQuA, and ARC-Challenge use a single-letter answer format, StrategyQA uses a yes/no format, and MathQA uses a worked-solution prompt that ends with ``The final answer is <letter>.'' The main table does not mix prompt recipes across methods within a benchmark.

\subsection{Compared Methods}

We compare seven families of runs.
\begin{enumerate}[leftmargin=1.5em]
\item Frozen base evaluation.
\item \full LoRA as a larger unconstrained reference.
\item \randomsub.
\item \pcasub.
\item \dora.
\item \pissa.
\item \mainmethod, the learned-basis warmup-and-freeze recipe.
\end{enumerate}

The main parameter counts are as follows. On Qwen, \full uses 967,680 trainable parameters, \randomsub and \pcasub use 972,352, \dora and \pissa use 682,496, and \mainmethod uses 682,596. On Llama, \full uses 991,232, \randomsub and \pcasub use 990,784, \dora and \pissa use 599,040, and \mainmethod uses 599,384. The most decision-relevant related-work comparison is therefore between \mainmethod and the budget-matched \dora/\pissa rows, while \full serves as a higher-budget unconstrained reference.

\subsection{Timing Study}

The trainable method is also evaluated in a separate switch-analysis setting with explicit warmup schedules. The main timing comparison studies a 100-step warmup followed by 300 constrained steps, and a 300-step warmup followed by 100 constrained steps. This ablation is secondary diagnostic evidence: it shows that an early basis can be representationally exact for the current weak adapter and still be poor for later continuation.

\subsection{Counterfactual Projection Test}

On GSM8K-partial we add a no-retraining intervention test on top of trained \full adapters from the main benchmark configuration. For each target module, we replace the learned write matrix by either its projection onto a candidate basis or the orthogonal residual left after that projection. We then re-evaluate the resulting model with no further optimization. This isolates write-space localization from training dynamics: if a candidate basis contains the useful signal, the projected-parallel model should retain most of the \full performance and the orthogonal residual should fail.

\subsection{Same-State Continuation and Top-k Continuation}

We also run a stronger continuation test. Starting from a fixed trained warmup checkpoint, we convert the same source state into constrained form under different bases and continue training with the same remaining budget and hyperparameters. For the learned basis itself, we then study concentration at two levels. First, we keep only the top-\(k\) learned directions inside each target module and continue training inside that smaller subspace. Second, we rank directions globally via single-direction ablations and measure whether globally selected top-\(M\) learned subsets outperform matched random subsets. These concentration analyses are reported on GSM8K, MathQA, and AQuA with seeds 39 and 40.

\subsection{Appendix-Only Controlled World}

The synthetic graph benchmark stays in the appendix. It isolates shortcut shift and structural length shift under controlled conditions and explains why write geometry can matter once the real-task diagnostics establish that the constrained parameterization itself is viable.

\section{Results}

\subsection{Write Geometry Is a Causal State Variable}

We begin with the strongest intervention in the paper. Table~\ref{tab:same-state-basis-swap} starts from the same warmup checkpoint, rewrites that identical trained state into constrained form under different bases, and continues training with the same remaining budget and hyperparameters. The learned same-task basis is best in every benchmark-backbone pair we test and beats the strongest control by more than 0.03 in five of the six rows. This establishes write geometry as a causal state variable: the same trained state has different futures under different write subspaces.

\begin{table}[t]
\centering
\small
\setlength{\tabcolsep}{4pt}
\resizebox{\linewidth}{!}{%
\begin{tabular}{@{}lrrrr@{}}
\toprule
Case & Learned & Random & PCA & Cross-task \\
\midrule
GSM8K / Qwen & \textbf{0.5703} & 0.3223 & 0.3203 & 0.3047 \\
GSM8K / Llama & \textbf{0.6348} & 0.5000 & 0.4883 & 0.4941 \\
MathQA / Qwen & \textbf{0.4980} & 0.1094 & 0.1152 & 0.1172 \\
MathQA / Llama & \textbf{0.4590} & 0.4023 & 0.3809 & 0.4258 \\
StrategyQA / Qwen & \textbf{0.6621} & 0.5645 & 0.5625 & 0.5625 \\
StrategyQA / Llama & \textbf{0.6895} & 0.6660 & 0.6719 & 0.6621 \\
\bottomrule
\end{tabular}%
}
\caption{Same-state basis-swap continuation from a fixed warmup checkpoint. For each case we start from the same trained \full state, convert it into constrained form under four different bases, and continue training with the same remaining budget and hyperparameters. Reported values are held-out test accuracy averaged over seeds 39 and 40. The learned same-task basis is best in every row and beats the strongest control by more than 0.03 in five of the six benchmark-backbone pairs.}
\label{tab:same-state-basis-swap}
\end{table}

The effect is sharpest on GSM8K/Qwen, where learned continuation reaches 0.5703 while the strongest control remains at 0.3223, and it remains substantial on GSM8K/Llama at 0.6348 versus 0.5000. MathQA shows the same pattern on both backbones. StrategyQA/Qwen still favors the learned basis by a wide margin, while StrategyQA/Llama shows the weakest separation and marks the current boundary of the claim. Once a useful state has formed, different subspace factorizations of that same state support different futures.

\FloatBarrier

\subsection{Counterfactual Projection Localizes the Useful Write Signal}

Table~\ref{tab:counterfactual-projection} turns basis overlap into a no-retraining intervention test. Starting from a trained GSM8K \full adapter, we replace each module-wise write matrix by either its projection onto a candidate basis or the orthogonal residual left after that projection, then re-evaluate the model with no further optimization. As a sanity check, projecting onto the final \full basis itself exactly reproduces \full on both backbones.

\begin{table}[t]
\centering
\small
\setlength{\tabcolsep}{5pt}
\resizebox{\linewidth}{!}{%
\begin{tabular}{@{}lrr@{}}
\toprule
\multicolumn{3}{l}{\textbf{Qwen2.5-3B-Instruct}} \\
\midrule
Basis family & Parallel kept & Orthogonal residual \\
\midrule
Learned basis & \textbf{0.5312} (95.9\%) & 0.2930 (4.1\%) \\
Random & 0.3125 (0.6\%) & \textbf{0.6172} (99.4\%) \\
Frozen-act PCA & 0.3105 (0.8\%) & 0.6113 (99.3\%) \\
\bottomrule
\end{tabular}%
}

\vspace{0.75em}

\resizebox{\linewidth}{!}{%
\begin{tabular}{@{}lrr@{}}
\toprule
\multicolumn{3}{l}{\textbf{Llama-3.2-3B-Instruct}} \\
\midrule
Basis family & Parallel kept & Orthogonal residual \\
\midrule
Learned basis & \textbf{0.6094} (95.2\%) & 0.4707 (4.8\%) \\
Random & 0.4766 (0.3\%) & 0.6328 (99.7\%) \\
Frozen-act PCA & 0.5078 (0.9\%) & \textbf{0.6387} (99.1\%) \\
\bottomrule
\end{tabular}%
}
\caption{No-retraining counterfactual projection test on GSM8K-partial. Starting from a trained \full adapter, we replace each module-wise write matrix by either its projection onto a candidate basis (parallel kept) or the complementary residual (orthogonal residual), then re-evaluate the model without further optimization. Parenthesized values report the corresponding fraction of \full write energy. The learned basis keeps most of the useful write signal in its parallel component, whereas random and frozen-activation PCA bases leave almost all useful signal in the orthogonal residual. Projecting onto the final \full basis itself exactly recovers \full and is used as a sanity check in the text rather than a separate row.}
\label{tab:counterfactual-projection}
\end{table}

The learned basis behaves qualitatively differently from random and frozen-activation PCA. On Qwen, its projected-parallel model keeps 95.9\% of the \full write energy and still reaches 0.5312 test accuracy, while the orthogonal residual keeps only 4.1\% of the energy and falls to 0.2930. On Llama, the same split is even sharper: the learned parallel component keeps 95.2\% of the write energy and reaches 0.6094, only 0.0098 below \full, whereas the orthogonal residual drops to 0.4707. The useful signal in the trained \full update is therefore localized inside the learned write space.

Random and frozen-activation PCA show the opposite pattern. Their projected-parallel models collapse to 0.3105--0.3125 on Qwen and 0.4766--0.5078 on Llama, while their orthogonal complements recover nearly all of \full. Useful write signal is concentrated inside the learned write subspace and lies almost entirely outside the guessed frozen bases.

Table~\ref{tab:counterfactual-transfer} extends the same counterfactual analysis to three additional cases.

\begin{table}[t]
\centering
\scriptsize
\setlength{\tabcolsep}{3.5pt}
\resizebox{\linewidth}{!}{%
\begin{tabular}{@{}lrrrr@{}}
\toprule
Case & Learned parallel & Learned orthogonal & Random parallel & PCA parallel \\
\midrule
MathQA / Qwen & \textbf{0.4746} (97.1\%) & 0.1035 (2.9\%) & 0.1309 (0.6\%) & 0.1172 (2.9\%) \\
MathQA / Llama & \textbf{0.4609} (99.5\%) & 0.4023 (0.5\%) & 0.3984 (0.4\%) & 0.3867 (1.9\%) \\
StrategyQA / Qwen & \textbf{0.6699} (97.6\%) & 0.5625 (2.4\%) & 0.5625 (0.5\%) & 0.5625 (1.2\%) \\
\bottomrule
\end{tabular}%
}
\caption{Additional counterfactual projection cases in the same mechanism analysis. Each row starts from a trained \full adapter on the stated benchmark, projects the learned write update onto the learned basis or onto a guessed frozen basis, and re-evaluates without further training. Parenthesized values report the retained fraction of \full write energy. Together with the GSM8K counterfactual table above, these rows show the same write-localization pattern: the learned basis keeps most of the useful behavior, while guessed frozen bases do not.}
\label{tab:counterfactual-transfer}
\end{table}

On MathQA/Qwen, the learned parallel component reaches 0.4746 while the orthogonal residual drops to 0.1035, and the random/PCA parallel controls remain at only 0.1309 and 0.1172. On MathQA/Llama, the learned parallel component reaches 0.4609, while the orthogonal residual drops to 0.4023 and the random/PCA parallel controls remain lower at 0.3984 and 0.3867. On StrategyQA/Qwen, the learned parallel component reaches 0.6699, while the orthogonal residual and both guessed-basis parallel controls all sit at 0.5625. Across all four cases, most of the useful \full write signal survives projection onto the learned basis, and very little survives projection onto guessed frozen bases.

\FloatBarrier

\subsection{Local Concentration Inside Modules}

The next question is how much of the learned basis is actually needed inside each module. Table~\ref{tab:topk-continuation} shows the answer. Across GSM8K, MathQA, and AQuA, the best per-module continuation always appears at \(k \in \{2,4\}\) across all twelve seed-level cases we test. \(k=8\) is never uniquely necessary.

\begin{table}[t]
\centering
\small
\setlength{\tabcolsep}{4pt}
\resizebox{\linewidth}{!}{%
\begin{tabular}{@{}lrrr@{}}
\toprule
Case & Top-2 & Top-4 & Top-8 \\
\midrule
GSM8K / Qwen & \textbf{0.5547} & 0.5488 & 0.5352 \\
GSM8K / Llama & 0.6230 & \textbf{0.6348} & 0.6328 \\
MathQA / Qwen & 0.5195 & \textbf{0.5273} & 0.5020 \\
MathQA / Llama & \textbf{0.4766} & 0.4609 & 0.4512 \\
AQuA / Qwen & 0.3543 & \textbf{0.3602} & 0.3543 \\
AQuA / Llama & \textbf{0.2539} & \textbf{0.2539} & \textbf{0.2539} \\
\bottomrule
\end{tabular}%
}
\caption{Same-state per-module top-\(k\) continuation under the learned basis. Starting from the same warmup checkpoint, we keep only the top-\(k\) learned directions inside each target module and continue training. Reported values are held-out test accuracy averaged over seeds 39 and 40. Across the twelve seed-level cases underlying this table, the best continuation is always achieved at \(k \in \{2,4\}\); \(k=8\) is never uniquely necessary.}
\label{tab:topk-continuation}
\end{table}

This is the clearest local concentration result in the paper. On GSM8K/Qwen, top-2 continuation reaches 0.5547, beating top-4 at 0.5488 and top-8 at 0.5352. On MathQA/Llama, top-2 reaches 0.4766, above top-4 at 0.4609 and top-8 at 0.4512. Even when top-4 is slightly better than top-2, as on GSM8K/Llama and MathQA/Qwen, the optimum still lies inside a very small per-module subset. Within modules, only a few learned directions are behaviorally effective.

\FloatBarrier

\subsection{Global Concentration and Sparse High-Impact Components}

Table~\ref{tab:global-topm} asks a stricter question. If we rank learned directions globally across modules rather than keeping top-\(k\) directions inside every module, does the concentration effect survive? The answer is yes, but at a larger budget. On GSM8K/Qwen, learned global top-32 reaches 0.5527 while a matched random subset reaches only 0.3203. On MathQA/Qwen, the same comparison is 0.5039 versus 0.1621. The gaps are smaller on AQuA and on some Llama rows, but the learned global subsets still dominate in most settings. Global concentration survives ranking, although the useful global budget is larger than the per-module top-2/top-4 budget.

\begin{table}[t]
\centering
\small
\setlength{\tabcolsep}{4pt}
\resizebox{\linewidth}{!}{%
\begin{tabular}{@{}lrrrr@{}}
\toprule
Case & Learned-4 & Random-4 & Learned-32 & Random-32 \\
\midrule
GSM8K / Qwen & \textbf{0.4414} & 0.3086 & \textbf{0.5527} & 0.3203 \\
GSM8K / Llama & \textbf{0.5137} & 0.4902 & \textbf{0.5859} & 0.4824 \\
MathQA / Qwen & \textbf{0.3359} & 0.1074 & \textbf{0.5039} & 0.1621 \\
MathQA / Llama & \textbf{0.4219} & 0.3926 & \textbf{0.4375} & 0.4199 \\
AQuA / Qwen & \textbf{0.3543} & 0.3248 & \textbf{0.3681} & 0.3445 \\
AQuA / Llama & \textbf{0.2165} & 0.2165 & \textbf{0.2539} & 0.2146 \\
\bottomrule
\end{tabular}%
}
\caption{Global top-\(M\) counterfactuals from single-direction ranking. We rank learned directions globally across modules, keep only the top-\(M\) learned directions, and compare against matched random subsets. Reported values are mean test accuracy over seeds 39 and 40. The learned subsets consistently outperform matched random subsets on the strongest arithmetic-style cases, showing that the concentration effect survives global selection, although at a larger scale than the per-module top-2/top-4 result.}
\label{tab:global-topm}
\end{table}

Table~\ref{tab:high-impact-components} adds the second structural piece: where the strongest components live.

\begin{table}[t]
\centering
\small
\setlength{\tabcolsep}{4pt}
\resizebox{\linewidth}{!}{%
\begin{tabular}{@{}l l r r l@{}}
\toprule
Case & Strongest component & Mean $\Delta$ & Flips & Dominant bias \\
\midrule
GSM8K / Qwen & L30 \code{q\_proj} d0 & -0.1426 & 48.5 & multistep / ratio \\
GSM8K / Llama & L20 \code{o\_proj} d0 & -0.0078 & 16.5 & multistep / ratio \\
MathQA / Qwen & L28 \code{down\_proj} d0 & -0.0957 & 52.0 & ratio \\
MathQA / Llama & L20 \code{down\_proj} d0 & -0.0312 & 48.5 & ratio \\
AQuA / Qwen & L28 \code{down\_proj} d0 & -0.0138 & 15.0 & ratio / multistep \\
AQuA / Llama & L20 \code{down\_proj} d0 & -0.0177 & 13.5 & add/sub / ratio \\
\bottomrule
\end{tabular}%
}
\caption{Highest-impact single directions from seed-pooled ablations. `Mean $\Delta$' is the mean test-accuracy change relative to the corresponding \full model when that one direction is removed. The strongest directions cluster in late \code{q\_proj}, \code{o\_proj}, and \code{down\_proj} modules rather than spreading uniformly across the adapter.}
\label{tab:high-impact-components}
\end{table}

Single-direction ablations identify a sparse set of late \code{q\_proj}, \code{o\_proj}, and \code{down\_proj} directions whose removal causes disproportionately large drops. The sharpest Qwen cases are late-layer \code{q\_proj} and \code{down\_proj} components; the strongest Llama cases shift across \code{o\_proj} and \code{down\_proj}, but still remain sparse and late. Additional rotated-basis controls are mixed, so the stable object is the sparse effect and its late-module localization rather than one privileged coordinate system inside the learned subspace.

\FloatBarrier

\subsection{Timing Is Secondary Diagnostic Evidence}

Table~\ref{tab:timing} provides supporting diagnostic evidence. It summarizes a timing study in the switch-analysis setting.

\begin{table}[t]
\centering
\small
\setlength{\tabcolsep}{5pt}
\resizebox{\linewidth}{!}{%
\begin{tabular}{lcccc}
\toprule
Warmup schedule & Qwen warmup \full & Qwen \mainmethod & Llama warmup \full & Llama \mainmethod \\
\midrule
100 + 300 & 0.3711 & 0.3145 & 0.5059 & 0.5508 \\
300 + 100 & 0.5156 & \textbf{0.5313} & 0.6016 & \textbf{0.6016} \\
\bottomrule
\end{tabular}%
}
\caption{Timing ablation for the trainable constrained recipe in the diagnostic switch-analysis setting. Each row reports held-out test accuracy averaged over seeds 39 and 40. Freezing the basis after only 100 warmup steps is too early, especially on Qwen. A 300-step warmup moves the method into a different regime: the constrained continuation improves over the weaker early-freeze schedule and preserves nearly all of the late-warmup performance. A separate Llama-only 350-step diagnostic run matched rather than improved the 300-step schedule and is discussed in the appendix.}
\label{tab:timing}
\end{table}

When the basis is frozen after only 100 warmup steps, the trainable constrained model is poor on Qwen and only modest on Llama. This is the clearest evidence that the write subspace can be exact for the current warmup adapter and still be the wrong subspace for the final task. A 300-step warmup changes the regime. On Qwen, the warmup \full model already reaches 51.56\% test accuracy in the diagnostic setting, and the constrained continuation improves it further to 53.13\%. On Llama, the same 300-step schedule produces a constrained model at 60.16\%, essentially matching the warmup model and closing most of the gap to the 60.94\% \full reference in that setting.

This timing result shows why a useful write space must first be formed. It does not identify which components inside that space are behaviorally essential. The same-state swap, projection, local top-\(k\), and global top-\(M\) results answer that stronger structural question.

\begin{figure}[t]
\centering
\includegraphics[width=\linewidth]{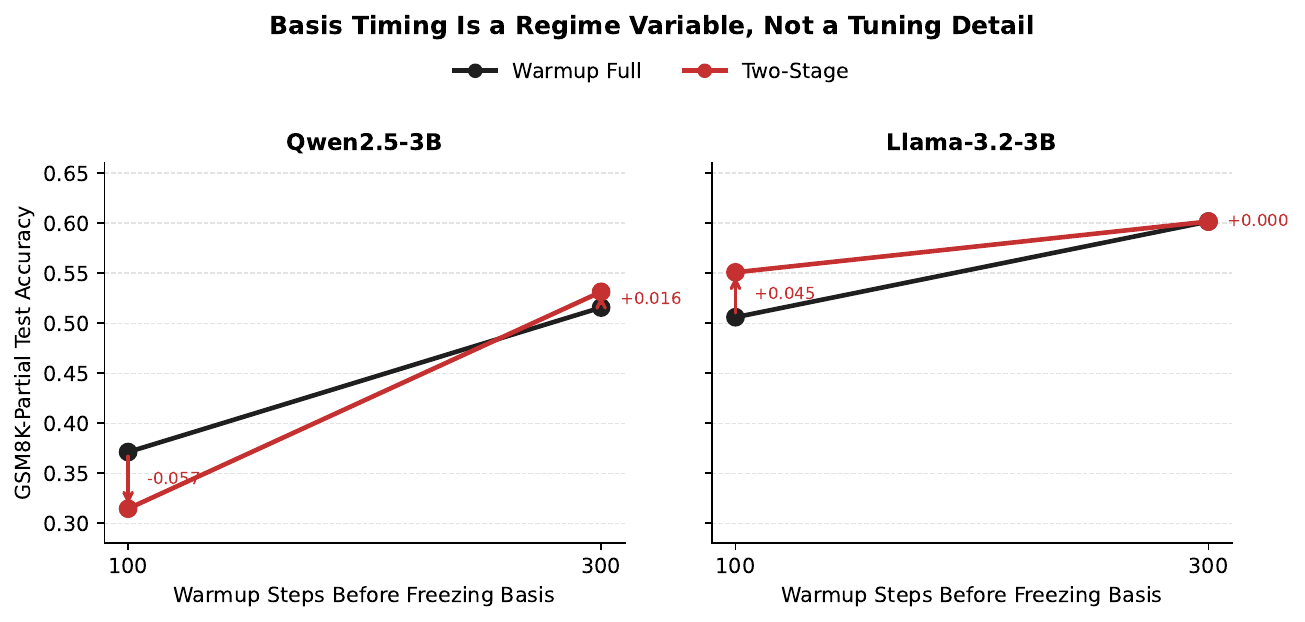}
\caption{Timing ablation for the trainable method. Freezing after 100 steps hurts because the learned write space is still immature. Freezing after 300 steps lands in a different regime: the constrained continuation improves over the warmup model on Qwen and preserves nearly all of it on Llama.}
\label{fig:timing-ablation}
\end{figure}

\FloatBarrier

\subsection{Main Real-Benchmark Table}

Table~\ref{tab:two-backbone} shows that the learned-basis recipe remains competitive beyond the mechanism probes. The table is split into a Qwen panel and a Llama panel so that backbone effects remain visible.

\begin{table}[t]
\centering
\small
\setlength{\tabcolsep}{3.5pt}
\resizebox{\linewidth}{!}{%
\begin{tabular}{lrrrrrr}
\toprule
\multicolumn{7}{l}{\textbf{Qwen2.5-3B-Instruct}} \\
\midrule
Method & GSM8K & CSQA & StrategyQA & AQuA & ARC-Challenge & MathQA \\
\midrule
Frozen base & 0.3242 & 0.7773 & 0.5625 & 0.3228 & \textbf{0.8242} & 0.1094 \\
\full & \textbf{0.6133} & 0.7813 & 0.6582 & \textbf{0.3720} & 0.8125 & \textbf{0.5000} \\
\randomsub & 0.3242 & 0.7754 & 0.6367 & 0.3346 & \textbf{0.8242} & 0.1191 \\
\pcasub & 0.2734 & 0.7734 & 0.6543 & 0.3386 & \textbf{0.8242} & 0.1758 \\
\dora & 0.6035 & 0.7773 & 0.6348 & 0.3484 & 0.8223 & 0.4727 \\
\pissa & 0.5664 & \textbf{0.7852} & 0.6367 & 0.3602 & 0.8125 & 0.4727 \\
\mainmethod & 0.5664 & 0.7793 & 0.6621 & 0.3445 & 0.8223 & 0.4980 \\
\bottomrule
\end{tabular}%
}

\vspace{0.75em}

\resizebox{\linewidth}{!}{%
\begin{tabular}{lrrrrrr}
\toprule
\multicolumn{7}{l}{\textbf{Llama-3.2-3B-Instruct}} \\
\midrule
Method & GSM8K & CSQA & StrategyQA & AQuA & ARC-Challenge & MathQA \\
\midrule
Frozen base & 0.4844 & 0.7188 & 0.6758 & 0.2441 & 0.7227 & 0.3711 \\
\full & 0.6191 & 0.7305 & 0.6680 & 0.2657 & 0.7109 & 0.4707 \\
\randomsub & 0.5020 & 0.7227 & 0.6699 & 0.2441 & \textbf{0.7246} & 0.4023 \\
\pcasub & 0.5020 & 0.7285 & 0.6484 & 0.2362 & 0.7188 & 0.4648 \\
\dora & 0.6348 & 0.7305 & 0.6797 & \textbf{0.2854} & 0.7207 & 0.4395 \\
\pissa & 0.6289 & \textbf{0.7383} & 0.6504 & 0.2500 & 0.7070 & 0.4629 \\
\mainmethod & \textbf{0.6367} & 0.7344 & \textbf{0.6855} & 0.2677 & 0.7227 & \textbf{0.4766} \\
\bottomrule
\end{tabular}%
}
\caption{Extended main real-benchmark table, reporting held-out test accuracy averaged over seeds 39 and 40. The table is split into two backbone-specific panels to keep backbone effects readable. \randomsub and \pcasub are fixed-before-training guessed-basis controls, \dora and \pissa are official PEFT baselines, and \mainmethod is the default learned-basis continuation row. GSM8K retains the prompt-aligned results used throughout the main benchmark.}
\label{tab:two-backbone}
\end{table}

The first pattern comes from GSM8K. On both backbones, the guessed frozen bases are weak: \randomsub and \pcasub sit far below the strongest tuned PEFT rows. On Qwen, \full is best at 0.6133, while \dora reaches 0.6035 and \mainmethod ties \pissa at 0.5664. On Llama, \mainmethod is the strongest row at 0.6367, narrowly above \dora at 0.6348 and \pissa at 0.6289. These cells match the mechanism picture above: once the learned write space has formed, a constrained continuation can stay strong.

The second pattern comes from the five additional benchmarks. On CommonsenseQA, \pissa remains strongest on both backbones, with \mainmethod staying close. On StrategyQA, \mainmethod stays strongest on both backbones. On AQuA, \full is strongest on Qwen and \dora is strongest on Llama. ARC-Challenge is near ceiling: on Qwen the strongest tied rows are Frozen base, \randomsub, and \pcasub at 0.8242, while \mainmethod reaches 0.8223 and matches \dora; on Llama, \randomsub is highest at 0.7246 and \mainmethod ties the base model at 0.7227. MathQA is more favorable: \full is strongest on Qwen at 0.5000 with \mainmethod close at 0.4980, while \mainmethod is strongest on Llama at 0.4766, above \full at 0.4707. The broad pattern is a competitive constrained recipe on several reasoning-heavy cells, while near-ceiling multiple-choice tasks remain less informative about the mechanism.

\subsection{What the Remaining Gap Means}

The two backbones support different parts of the story. Llama is the cleanest end-to-end case for \mainmethod in the main table: it leads on GSM8K, both StrategyQA cells, and MathQA, and it stays near the top on CommonsenseQA and ARC-Challenge. Qwen is more mixed: \mainmethod leads on StrategyQA, stays within 0.0020 of \full on MathQA, and reaches 0.8223 on ARC-Challenge, while \full remains strongest on GSM8K and AQuA. The remaining challenge is to learn and freeze the basis robustly enough that the trainable recipe wins more consistently on the hardest Qwen cells.

\subsection{What the Appendix Adds}

The appendix serves two purposes. First, it records the per-seed real-benchmark numbers so that the timing story is auditable. Second, it preserves the synthetic graph evidence in the right role. In that controlled world, constrained write spaces improve shortcut-shift and length-shift generalization much more cleanly than they do on GSM8K. Those gains explain why write geometry is worth studying in the first place.

\section{Discussion}

The paper's main result is simple: LoRA does not use its write budget uniformly. A small, structured subset of write components carries most of the behavior. The evidence is consistent across several intervention levels: same-state continuation shows that write geometry is causal, projection shows that useful signal is localized to the learned write space, local and global concentration tests show that the effect is sharply non-uniform, and single-direction ablations show that the strongest components cluster in a sparse set of late write modules.

This changes how to interpret write geometry in PEFT. The central question is how much of the learned write update is actually needed for behavior. On the current arithmetic-style reasoning family, the useful update is substantially smaller and more structured than the full trained low-rank write, with especially strong concentration inside modules and a weaker but still real concentration at the global level.

This perspective also clarifies what the smaller parameter counts mean. The parameter savings relative to the larger \full reference are real: \mainmethod uses 29.5\% fewer trainable parameters on Qwen and 39.5\% fewer on Llama. More importantly, the results show that rank budget and behavioral effectiveness are different objects. Recent post-hoc compression and spectrum-editing methods already suggest that trained adapters contain internal redundancy \citep{vulic2026lorasqueeze,tian2026spectralsurgery}. Our interventions make that point behaviorally explicit and localize redundancy at the level of write directions. Structured or routed adapter systems can be read through the same lens: they shape where the model is allowed to write, combine, or preserve task signal \citep{xiao2026structlora,chang2026soslora,wu2024mole,li2024mixlora,zhu2026fcprag}.

\subsection{Limitations}

The paper does not claim a unique canonical basis-column mechanism. Rotated controls are mixed, which indicates that the stable object is the sparse structured write effect rather than one privileged coordinate system inside the learned subspace. The global concentration effect is also weaker than the per-module concentration effect, so the current evidence does not reduce the whole model to only two or four global directions. Finally, this is a structural mechanism paper rather than a full semantic explanation paper. We identify sparse high-impact write components and where they live, but we do not explain every component in natural-language algorithmic terms.

\section{Conclusion}

This paper isolates the geometry of LoRA writes after adaptation and shows that the behaviorally effective part of a trained update is sparse and structured. Projection shows that useful signal is localized to the learned write space. Same-state continuation shows that different write subspaces give the same trained checkpoint different futures. Per-module top-\(k\) continuation shows strong local concentration, global top-\(M\) selection shows that this concentration survives global ranking, and single-direction ablations show that a sparse set of late \code{q\_proj}, \code{o\_proj}, and \code{down\_proj} components carries outsized behavioral impact.

The contribution is a structural mechanism result with practical consequences. Geometry-constrained low-rank adaptation remains viable on real tasks, and the stronger finding is how little of the learned write space is actually needed. Future PEFT methods should allocate budget toward high-impact write components instead of treating every direction in a low-rank update as equally valuable. \mainmethod is useful because it exposes that internal structure clearly enough to measure it.

\bibliographystyle{unsrtnat}
\bibliography{references}

\clearpage
\appendix
\section{Additional Real-Benchmark Details}

This appendix records the diagnostic real-benchmark runs that underlie the mechanism sections in the main text. These runs use train/validation/test sizes of 1,024/128/256, seeds 39 and 40, and a shared target-module family over the last four layers. They should be read as mechanism diagnostics rather than as the main benchmark table.

The appendix now has two jobs. First, it preserves the timing diagnostics that explain why a useful learned write space exists. Second, it records the continuation diagnostics that show how little of that write space is actually needed once it has formed.

\subsection{Timing Diagnostics}

\begin{table}[t]
\centering
\small
\setlength{\tabcolsep}{5pt}
\resizebox{\linewidth}{!}{%
\begin{tabular}{crrrrrr}
\toprule
Seed & Base & \full & \randomsub & \pcasub & Warmup \full & \mainmethod \\
\midrule
39 & 0.2266 & 0.4688 & 0.2813 & 0.2773 & 0.5156 & 0.5430 \\
40 & 0.2266 & 0.5469 & 0.1992 & 0.2813 & 0.5156 & 0.5195 \\
\bottomrule
\end{tabular}%
}
\caption{Per-seed GSM8K-partial test accuracy for the diagnostic Qwen2.5-3B-Instruct timing study. ``Warmup \full'' refers to the 300-step unconstrained warmup used by \mainmethod before basis freezing. The key pattern is stable in this diagnostic setting: both seeds end with constrained models above the corresponding \full baseline.}
\label{tab:qwen-seeds}
\end{table}

\begin{table}[t]
\centering
\small
\setlength{\tabcolsep}{5pt}
\resizebox{\linewidth}{!}{%
\begin{tabular}{crrrrrr}
\toprule
Seed & Base & \full & \randomsub & \pcasub & Warmup \full & \mainmethod \\
\midrule
39 & 0.4570 & 0.6055 & 0.5039 & 0.4648 & 0.5820 & 0.5781 \\
40 & 0.4570 & 0.6133 & 0.4531 & 0.4883 & 0.6211 & 0.6250 \\
\bottomrule
\end{tabular}%
}
\caption{Per-seed GSM8K-partial test accuracy for the diagnostic Llama-3.2-3B-Instruct timing study. The 300-step warmup gives one seed slightly above \full and one seed slightly below it, which explains why the mean trainable result is a near-match in this diagnostic setting.}
\label{tab:llama-seeds}
\end{table}

In this diagnostic setting, Qwen is the clearest end-to-end success case and Llama is more conservative, but both show the same high-level pattern: an early basis can be exact for the current weak adapter and still be poor for later continuation, while a later learned basis is much stronger. These timing records explain why the learned write space should be treated as a meaningful object in the first place.

\subsection{Same-State Basis-Swap Diagnostics}

The main text reports seed-averaged same-state basis-swap continuation results across GSM8K, MathQA, and StrategyQA. The per-seed records matter because they show that the learned same-task basis is consistently above the strongest guessed-basis control in nearly every arithmetic-style case. The weakest separation appears on StrategyQA/Llama, which is why the manuscript treats that pair as a boundary case.

\subsection{Top-k Continuation Diagnostics}

The strongest new mechanism result is the same-state top-\(k\) continuation analysis. Starting from the same learned-basis warmup checkpoint, we retain only the top-\(k\) learned directions and continue training. The main text reports the seed-averaged results on GSM8K, MathQA, and AQuA. The corresponding per-seed records show two robust facts. Best continuation does not require the full learned subspace, and the optimum appears in a very small range, typically top-2 or top-4, instead of increasing monotonically with larger \(k\).

\section{Synthetic Graph Evidence as Mechanism Support}

The synthetic graph world remains valuable because it exposes two failure modes that are difficult to isolate cleanly in a real benchmark: shortcut-sensitive surface cues and structural length shift. We keep it in the appendix because the paper's main claim is anchored in real-task evidence.

\begin{table}[t]
\centering
\small
\setlength{\tabcolsep}{5pt}
\resizebox{\linewidth}{!}{%
\begin{tabular}{@{}lrrrr@{}}
\toprule
Method & Trainable params & Val & Shortcut shift & Length shift \\
\midrule
\full ($r=16$) & 1,359,872 & 0.8281 & 0.8444 & 0.6764 \\
\full (budget-match) & 967,680 & 0.8815 & 0.8965 & 0.7220 \\
\randomsub & 967,424 & 0.8854 & 0.9043 & 0.7533 \\
\probesub & 967,424 & 0.8958 & 0.9043 & 0.7546 \\
\pcasub & 967,424 & \textbf{0.9036} & \textbf{0.9147} & \textbf{0.7624} \\
\bottomrule
\end{tabular}%
}
\caption{Appendix-only synthetic graph summary in the controlled shortcut/length-shift world. A fixed write bottleneck helps most strongly when shortcut shift and structural length shift can be separated cleanly.}
\label{tab:synthetic-graph}
\end{table}

The pattern in Table~\ref{tab:synthetic-graph} is substantially stronger than the GSM8K result. Under controlled shortcut and length shifts, all constrained variants beat the fixed-rank \full baseline, and \pcasub also beats the near-budget \full control on all three held-out metrics. The synthetic world therefore acts as a microscope for the mechanism: when task structure and shortcut-sensitive directions can be separated cleanly, constraining writes is a strong inductive bias. On real benchmarks, the same bias still helps, but the gains are smaller because the separation is noisier and the update needs are broader.

\end{document}